# Semantic Modeling with SUMO

Robert B. Allen

[0000-0002-4059-2587]

rba@boballen.info

**Abstract:** We explore using the Suggested Upper Merged Ontology (SUMO) to develop a semantic simulation. We provide two proof-of-concept demonstrations modeling transitions in a simulated gasoline engine using a general-purpose programming language. Rather than focusing on computationally highly intensive techniques, we explore a less computationally intensive approach related to familiar software engineering testing procedures. In addition, we propose structured representations of terms based on linguistic approaches to lexicography.

**Keywords:** Definitions, Description Logic, Model-Checking, Model-Level, Rules, Semantic Simulation, Transitionals, Truth Maintenance

## 1  Introduction

We believe knowledge representation should be fully integrated with programming languages. Therefore, we are exploring the implementation of dynamic semantic simulations based on ontologies using a general-purpose programming language (cf., [4]). These simulations allow model-level constructs such as flows, states, transitions, microworlds, generalizations, and causation, and language features such as conditionals, threads, and looping.

In this paper, we provide initial demonstrations for how the Suggested Upper Merged Ontology (SUMO) can be applied to Python-based semantic modeling. SUMO has both a rich ontology and a sophisticated inference environment built to use first-order predicate calculus [9, 15, 16, 25, 27, 28].[1] The SUMO ontology incorporates approaches from several other ontologies ([28] p94). Like the Descriptive Ontology for Linguistic and Cognitive Engineering (DOLCE) [12], SUMO also incorporates insights from linguistics. In fact, one extension of SUMO explores Natural Language Generation ([17]).

The SUMO ontology is implemented with SUO-KIF, which is a subset of KIF (Knowledge Interchange Format) [18]. KIF is a notation based on the operators of first-order logic (FOL).[2] As a type of description logic, SUMO includes rules, which are implemented with formulas. These represent constraints about the world.

Computationally intensive theorem proving for large ontologies has been the focus of much of the recent research on SUMO. By comparison, we explore state-based modeling for small example applications. This is a companion to [4] semantic modeling using the Basic Formal Ontology (BFO) [10]. As in that work, the interactions studied are object-driven. We do not focus on complex inference in this paper; rather we apply simple test cases analogous to those used in requirements testing and model checking [14, 18] to detect possible conflicts in domains, states, and relationships following Transitions.

Truth maintenance [24] considers how to ensure that there are only true statements in a knowledgebase as new statements are added. The original knowledgebase is assumed to be true and any incoming statements that conflict with those are rejected. Research on Truth Maintenance Systems (TMS) explores robust and general abstractions to detect and resolve conflicts. In the interest of practical applications, we support lightweight, tractable approaches to inference and truth maintenance. These approaches are related to those from software engineering used to develop test cases. We do not attempt automated repair of the truth values. Rather, inconsistencies are highlighted for the developer. In short, our approach is related to TMS but more modest.

---

[1] General information about the SUMO project is available at http://ontologyportal.com/ap/. SUMO's full KIF files, its code and other tools are at https://github.com/ontologyportal/sumo. The bulk of the SUMO ontology and the software are available with only light restrictions. When first exploring SUMO, we found it best to use only a few of the KIF files, together with the Python interface, and a moderately powerful computer.

[2] FOL has been criticized as allowing too much flexibility [26, 33] to be a suitable platform for ontologies. However, much of the SUMO ontology avoids the possible pitfalls and any lapses could be remedied. Indeed, we believe that BFO itself could be largely implemented in SUO-KIF.

## 2    Details about SUMO

All terms in SUMO are descended from Entity. One branch, Abstract Entities, includes Relations, Predicates, Subclasses, and Instances. Abstract Entities form the Structural Ontology, which allows us to construct axioms (and rules, as described below) in SUO-KIF. An example of a typical axiom is: (instance thisGasEngine GasolineEngine). This is read as "thisGasEngine is an instance of the class GasolineEngine". "instance" is a BinaryPredicate that relates the instance to the class.[3]

SUMO's main ontology file, Merge.kif, contains the Structural Ontology and Base Ontology. The SUMO GitHub distribution also has about 25 other .kif files including the Mid-Level-Ontology.kif (MILO) and the North American Industrial Classification System as naics.kif.[4] Taken together, these files can be considered a type of foundry. Rules are structured using the KIF operators: implies or IF, (=>, Exists, and And.[5] Other KIF operators include Or, ForAll, Not, and IF-and-only-IF, (<=>.

In Dining.kif, one of the lower-level ontologies, we find "(subclass Bakery Business)". The SUMO term Bakery is also associated with the rule shown in Figure 1. Query terms are preceded by a "?" so "(instance ?BAKERY Bakery)" is read as "is there an instance of a Bakery". The rule asserts that a Bakery does Baking of Food for Humans and that it is a CommercialService.

| (=> | IF |
|---|---|
| (instance ?BAKERY Bakery) | there is an instance of a BAKERY |
| (exists (?SERVICE ?FOOD ?BAKE) | THEN |
|  | there exists a Service that Bakes Food |
| (and | AND |
| (instance ?BAKE Baking) | there is an instance of Baking |
| (result ?BAKE ?FOOD) | that Baking is baking of Food |
| (instance ?FOOD (FoodForFn Human)) | that Food is Food for Human consumption |
| (agent ?BAKE ?BAKERY) | the Bakery does the Baking |
| (instance ?SERVICE CommercialService) | there is a CommercialService |
| (agent ?SERVICE ?BAKERY) | the Bakery is the agent of that CommercialService |
| (instance ?SERVICE Selling) | the CommercialService engages in Selling |
| (patient ?SERVICE ?FOOD)))) | this CommercialService sells the Food the Bakery Bakes |

*Figure 1:* Rule associated with a Bakery from Dining.kif. Interpretation is added on the right.

We are interested in the relationship between knowledge representation and programming languages. While the SUMO ontology allows states and state changes, states in themselves are not distinct, "real" entities. Thus, they are not accepted into so-called realist ontologies such as BFO.

Rules are also integral to the definition of Processes in SUMO. For instance, the Process of Cooking is considered to be the preparation of food and Baking is Cooking using an Oven. Figure 2 shows a SUMO Rule for the Process of TurningOffDevice, which is a type of InternalChange. DeviceOn is an instance of an internal state that may be changed. Another example of StateChange in the SUMO ontology is an explicit PhysicalState Attribute (i.e., Phase Changes) for Substances whose change is described with an explicit StateChange Process.

---

[3] By convention, instances are usually identified with lower case letters. Care is sometimes required to interpret the statements. For instance, "attribute" is a predicate while "Attribute" is an Abstract Entity.

[4] The distribution also includes YAGO-SUMO ([16], [28] chp9) that has many facts pulled from Wikipedia. In addition, WordNet terms ([28] chp5) are mapped to SUMO terms.

[5] In some cases (e.g., Figure 1), rules specify context and attributes for Objects. In other cases (e.g., Figure 2) they can specify Objects associated with processes (e.g., Figure 2). In addition, rules are used in the Structural Ontology to specify the effects and constraints of Relations such as Predicates.

| | |
|---|---|
| (=> | IF |
| (and | AND |
| (instance ?P TurningOffDevice) | there is an instance of TurningOffDevice |
| (patient ?P ?D)) | the patient of TurningOffDevice is a Device, D |
| | THEN |
| (and | AND |
| (holdsDuring | begins during |
| (BeginFn (WhenFn ?P)) | some interval |
| (attribute ?D DeviceOn)) | when the device is on |
| (holdsDuring | ends during |
| (EndFn (WhenFn ?P)) | some interval |
| (attribute ?D DeviceOff)))) | when the device is off |

*Figure 2.* Rule associated with the Process of TurningOffDevice from Merge.kif. Interpretation is added on the right.

## 3   Semantic Modeling Using a Simplified Python-Based Version of SUMO

### 3.1   A Simplified Python SUMO Environment

In the SUMO software, SUO-KIF statements are processed by KB.java. KB.java is descended from software originally developed by Teknowledge Inc. It supports theorem proving engines such as Vampire and E. Much of the recent research with SUMO has emphasized Typed First-Order Logic.

We implemented a portion of the SUMO software functionality in Python. Python was selected because it supports a variety of data structures and direct access to internal class structures. As described in Sections 4 and 6, we applied the program to a fragment of the SUMO ontology.[6] Our primary goal was a demonstration of the concept.

### 3.2   Semantic Modeling with Simplified Python SUMO

The primary goal of ontologies is to describe the types of objects in the world and the types of relationships among those objects. There is less attention to how simple objects (e.g., parts) are combined into complex objects such as Systems and Devices and how Systems would participate in simulations.

**Transitions, States, and Mechanisms:** We identify Transitions (i.e., actual state changes) as a distinct type of Process. Sections 4 and 6 provide examples of applications that use Transitions. Flows are sequences of Transitions. In previous work, we have discussed several types of Flows such as Mechanisms, Workflows, and ad hoc Causal Sequences. Note that as shown in Figure 2, a variety of conditions can be imposed on a Transition. For instance, Transitions may be restricted to occur only in certain States. While object-oriented models are not necessarily used to develop state machines, they can be applied that way [11].

Cua, et al. [15] describe story plans with a SUMO-based knowledge representation of a narrative. This is similar to the development of scenarios such as we are proposing. However, Cua et al. stopped short of executing the plans it developed.

**Parts, Inputs and Outputs:** Potentially, structured descriptions could be developed for complex objects [4, 13, 31, 35] such as Systems and Devices. These complex objects are collections of interacting Mechanisms and Parts [3], which may be described with some combination of behavior, structure, and functionality. Because a System or Device is a unit, we know that its parts move with it as it moves in the microworld. Systems also have inputs and outputs. For example, gasoline is required for the operation of a gasoline Engine but is not part of the Engine; it is an input; exhaust is an output.

---

[6] SUO-KIF uses different terminology from object-oriented systems. For instance, a SUMO class is part of the Structural Ontology rather than a programming language object. We implemented the instances of elements of the Structural Ontology as Python classes while recognizing the distinction between types of classes. Instances associated with those elements were built directly with the Python code associated with the class.

**Rules and Microworlds:** In our Python version of SUMO, rules were implemented as methods associated with classes. We take rules, such as the description of the Bakery in Figure 1, as making assertions about Entities that exist in a microworld. A microworld provides a frame (i.e., context or scope) for all instantiated objects in the simulation. In our models, the microworld is a spatial region in which objects are located. Potentially, the microworld could have properties of its own (e.g., gravity, air). Moreover, the microworld may be divided in spatial-temporal sub-regions.

Because of the rule in Figure 1, we know that a Bakery must sell edible baked goods. If Bakeries do Baking, we assume that they have Ovens because an additional Rule associated with Baking requires that there be one. However, the SUMO rules do not always provide enough information about the implemented models. Future work is needed to support greater consistency.

**Validation:** Because our focus is on description more than inference, we did not implement advanced inference techniques such as backtracking. Rather, we focused on whether the network had conflicts and Relationships such as transitions were applied.[7] As with many systems of linked data, not all Relationships are expressed explicitly; some are implied by other relationships. Thus, we activated Relationships and applied any associated rules to build a set of inferred relationships. We then examined for conflicts such as connections of logically contradictory attributes such as a device being both On and Off simultaneously, or connections that are inconsistent with the model such as Combustion occurring with Exhaust gasses.[8]

Because we use run-time validations as guard conditions or contracts for transitions, and the transitions change the active entities and relationships, the networks of inference relationships may need to be reconstructed for every transition. This is cumbersome, but the burden may be reduced. If a transition is repeating in an otherwise unchanging environment, it does not have to be revalidated in each cycle.

## 4    Example: Engine Ignition Control Switch

The first example describes a simple transition and its validation. In this example, an Engine is alternately turned On and Off. It follows the rule in Figure 2, implemented in Python code with one time interval (temporally) following another in the simulation. TurningOn the engine breaks the relationship between the Engine and the EngineOff attribute and connects the EngineOn state to the Engine (cf., [6]). The active relationships are managed with Python dictionaries.[9] The transition is validated in several ways (see Section 3.2). First, we extend the explicitly connected relationships with inferred relationships and check them for conflicts. Then we apply probes to check for specific conflicts. For instance, we check that only one Relationship is connected to EngineState as required by the partition of EngineOn and EngineOff attribute instances. Other constraint probes could examine conflicts in location or behavior.

## 5    Definitions

Terms in ontologies are often accompanied by definitions. However, those definitions are often ad hoc; it would seem desirable for definitions to be well-structured and expressed by the terms in the ontology. [4] proposed a definition for a waterfall as a programmatic simulation based on ontology terms. It described the interaction of entities implicitly associated with a waterfall such as water and a streambed. A definition such as that in [4] could be compiled directly into the Python code for the simulations, to be applied whenever a waterfall object was instantiated. Rules in the SUMO ontology may be considered definitions. In fact, the SUMO ontology includes text documentation for each term that mirrors the axioms and rules.

Lexical semantics describes how meaning is assigned to words. It incorporates insights from computational linguistics, drawing from [23], which categorizes verbs according to their alternations, and from [18], which proposed a "generative semantics" based on semantic primitives. For example, SUMO supports CaseRoles such as agent and patient. [1] proposed using Fillmore's [32] frames. Following DOLCE, BFO [19] applied Vendler's analysis of the aspect of verbs [34]. All these approaches are considered as part of the Event Structure of the verbs.

---

[7] We also checked the validity of the domains of entities as they were instantiated into the model.

[8] A detailed model could determine that Combustion cannot occur without sufficient oxygen but we can apply a higher-level check without such detailed modeling.

[9] Larger networks could use JSON documents or even link to a NOSQL database such as MongoDB.

Another approach to lexical semantics proposes a "generative lexicon", in which meaning is assigned to words by compositionality, that is by a nuanced balancing of several factors [29, 30]. Specifically, it proposes that there are four levels of semantic representation (Figure 3). The Argument Structure is based on syntactic categories (e.g., direct objects). The event Structure is as described above. The Qualia are subdivided by Aristotle's four types of causation (*aitia*). Lexical Inheritance is the relationship to other words in the corpus.

| Argument Structure | Specification of number and type of logical arguments and how they are realized syntactically. |
|---|---|
| Event Structure | Definition of the event type of a lexical item and a phrase. Sorts include STATE, PROCESS, and TRANSITION, and events may have a subevent structure. |
| Qualia | Modes of explanation. |
|     Formal | That which distinguishes it within a larger domain |
|     Telic | Its purpose and function[10] |
|     Constitutive | The relationship between an object and its constitutive parts |
|     Agentive | Factors involved in its origin or "bringing it about" |
| Lexical Inheritance | Identification of how a lexical structure is related to other structures in the type lattice, and its contribution to the global organization of a lexicon. |

*Figure 3:* Factors associated with the generative lexicon. Comments on the right are from ([29] p61 and p76).

The four dimensions of the Qualia could be the basis for a structured dictionary schema to be used for structured knowledge. Although they are developed for natural lexicons, these dimensions should also be useful for constructing ontologies. While existing ontologies include some features of the Qualia, those are typically ad hoc rather than systematically structured.[11] Further, the Qualia could, potentially, be incorporated into an ontology using rules such as those in Figures 1 and 2. In any event, any collection of definitions, rules, and ontologies should be more systematically developed and cross-referenced than is done in any resource currently available.

We also envision what could be called a hyper-partonomy. That would be a top-down graph for complex objects which combined a rich ontology with a partonomy. As specific components are identified, the space of possible objects could be progressively constrained. For instance, if we knew that a given object included a steering wheel, we could conclude that it is a type of motor vehicle. If we then learned that it has a spark plug, we could narrow the possibilities to gasoline-powered motor vehicles, and so on. Moreover, the domains of associated Processes can be progressively refined.

## 6 Example: 4-Stroke Engine with Limited Fuel

As a second example implementation, we adapted and simplified the description of a 4-stroke gasoline engine from Cars.kif. The engine is a type of system; the pistons may be considered its power sub-system. A 4-stroke gasoline engine is highly predictable. We model one piston with transitions through the four strokes (Intake, Compression, Combustion, and Exhaust). We did not model the electrical subsystem.

The piston, valves, fuel, spark plug, and crankshaft are included in the model. As the simulation of the engine runs, fuel is consumed.[12] In each cycle, the valves allow gasoline to enter and exhaust to exit the piston. After each transition, the relationships among the components of the simulated piston are validated with procedures similar to those described in Section 4. The ignition switch (as described in Section 4) and piston run as separate threads. When the simulated piston runs out of fuel, the engine stops. The engine can also be halted if the ignition switch is turned off. Thus, multiple conditions are checked at the beginning of each cycle (cf., [3]). We supported focused testing of several other features. In particular, we supported tests for each transition such as checking that the piston is in the compress position before the spark.

---

[10] The telic dimension is further broken down as direct or indirect ([29] p99-100). Direct telic seems roughly analogous to Dispositions.

[11] [22] explored mining SUMO rules to enhance the Rich Event Ontology. We believe that it could be more effective to incorporate the Rich Event Ontology as a component of the SUMO ontology.

[12] In several cases, we grouped parts together in Assemblies [4]. For instance, the Piston was composed of Piston head and Piston rod.

## 7 Discussion

### 7.1 The Model Layer

Semantic models generalize instances. They may include Model-level descriptions with constructs such as states that are not "real", although we implicitly assert that those Model-level descriptors potentially could be fully explained within the model framework.

Complex objects such as an engine could have "metadata" wrappers. As specified in MILO, an Engine is a Transducer and a Transducer "is a device which is capable of converting one form of energy into another". Thus, descriptions reflect not only the interaction of its parts but also its inputs and outputs.[13] We could also support the description of cross-granular processes such as the description of Combustion at both the molecular level and the higher-level effect of pushing the Piston down. Finally, to the extent these models are state-based, techniques such as temporal model checking can be applied [14, 19].

### 7.2 Extending the SUMO Ontology

While the SUMO ontology has broad coverage, there are many additional areas where it might be productively expanded. Here, we consider briefly digital humanities and scientific research reports. Other areas could include medicine, geology, and criminal justice.

**Digital Humanities:** YAGO-SUMO includes many facts about artists and painting but not much about the context of their lives and work [16]. If the Getty Research Institute's Art and Architecture Thesaurus (AAT) were incorporated into SUMO, it could help support the implementation of that context [8].

Just as narrative scenarios can be developed for stories (e.g., [15]), we propose that they can be applied to the description of history. For instance, we could provide a structured "community model" for the GangJin pottery village in 12[th] century Korea (Goryeo Dynasty) [7]. Such structured descriptions could be applied across all areas of digital history, even if there is less predictability about the causal relationships among the parts of a community than there is for an engine.

**Research Reports and Data Sets:** We have proposed that direct representation of scientific research reports replace traditional text-based research reports [2]. Scientific propositions are similar to logical axioms. Indeed, the goal of science is to develop a consistent and efficient set of axioms and rules that cover natural phenomena. Supporting evidence is observations that are shown to be consistent with the model.

We should be able to develop rich knowledge management tools that could simulate the interaction of entities and even to "directly represent" entire research reports [3]. A broad ontology such as SUMO should be particularly helpful for organizing social science research data [5] and describing the high-level issues addressed in the introductions and conclusions of the research reports.

### 7.3 Conclusion

SUMO has an extensive ontology and is an impressive basis for research on inference for large knowledgebases. In this paper, we have explored its potential for smaller-scale applications that do not focus primarily on complex inference. We have demonstrated some initial steps to using the SUMO ontology as the basis for model-level descriptions which could then be applied to the description of instances. We believe that by combining insights from simulation, knowledge representation, programming languages, argumentation, and linguistics, we can build a new generation of information management tools.

---

[13] In the case of Engines, this could include outputs such as Exhaust (see Section 5) as well as the rotation of the crankshaft.